\documentclass[runningheads]{llncs}
\usepackage{graphicx}
\usepackage{amsmath,amssymb} % define this before the line numbering.
\usepackage{color}
\usepackage{multirow}
\usepackage{amsmath}
\usepackage{amsfonts}
\usepackage{algorithm}
\usepackage{algorithmic}
\usepackage{multirow}
\usepackage{colortbl}
\usepackage{color}
\usepackage[table]{xcolor}
\usepackage{epigraph}
\usepackage{graphicx}
\usepackage{caption}
\usepackage{subfigure}
\usepackage{hyperref}
\usepackage{float}
\usepackage{longtable}
\usepackage{pdfpages}
\usepackage{pdflscape}
\usepackage[acronym,toc]{glossaries}
\usepackage{setspace}
\usepackage{wrapfig}
\usepackage{amssymb}
\usepackage{stmaryrd}
\usepackage{array}
\usepackage{enumerate}
\usepackage{paralist}
\usepackage{tabularx}
\usepackage{epsfig}
\usepackage{array}

\def\ie{\emph{i.e.}}

\def\etal{\emph{et~al.}}
\newcommand\norm[1]{\left\lVert#1\right\rVert}

\begin{document}
\pagestyle{headings}
\mainmatter

\def\ACCV20SubNumber{***}  % Insert your submission number here

%===========================================================
\title{Efficient Depth Completion Using Learned Bases} % Replace with your title
\titlerunning{PCA Depth}
% If the paper title is too long for the running head, you can set
% an abbreviated paper title here
%

%
\authorrunning{Zhong et al.}
% First names are abbreviated in the running head.
% If there are more than two authors, 'et al.' is used.
%
\author{Yiran Zhong$^{1}$, Yuchao Dai$^{2}$, Hongdong Li$^{1}$}
\institute{$^{1}$Australian National University, 
$^{2}$Northwestern Polytechnical University\\
\tt\small{\{yiran.zhong, hongdong.li\}@anu.edu.au}, daiyuchao@nwpu.edu.cn}

\graphicspath{{Figures/}}
\maketitle

%===========================================================
\begin{abstract}
In this paper, we propose a new global geometry constraint for depth completion. By assuming depth maps often lay on low dimensional subspaces, a dense depth map can be approximated by a weighted sum of full-resolution principal depth bases. The principal components of depth fields can be learned from natural depth maps. The given sparse depth points are served as a data term to constrain the weighting process. When the input depth points are too sparse, the recovered dense depth maps are often over smoothed. To address this issue, we add a colour-guided auto-regression model as another regularization term. It assumes the reconstructed depth maps should share the same nonlocal similarity in the accompanying colour image. Our colour-guided PCA depth completion method has closed-form solutions, thus can be efficiently solved and is significantly more accurate than PCA only method. Extensive experiments on KITTI and Middlebury datasets demonstrate the superior performance of our proposed method. 
\end{abstract}

%===========================================================

\section{Introduction}
Acquiring dense and accurate depth measurements is crucial for various applications such as autonomous driving \cite{Chencvpr2016}, indoor navigation \cite{BiswasV12}, robot SLAM\cite{Yi2017}, virtual or augmented reality \cite{Yang2014690}. However, due to the limitation of current depth sensing technology, captured depth maps are often in a sparse form (\ie LiDAR) or suffering severe data missing problem on transparent and reflective surfaces (\ie Microsoft Kinect, Intel RealSense, and Google Tango). How to effectively interpolate these sparse depth points becomes an active topic recently.

Depending on the input modalities, the depth completion task can be subdivided to depth inpainting, depth super-resolution and depth reconstruction from sparse samples. Lots of works have been proposed to address this problem. For conventional methods, researchers often employ Markov Random Fields (MRF) to interpolate sparse depth points by encouraging nearby points to have similar depth values. The main drawback is that these methods often require several seconds to minutes to process a single frame and may create lots of artifacts on recovered depth maps. Deep learning based methods can produce a dense depth map in real time but it often requires lots of training data and suffers generalization issues. 

In this paper, we propose a general geometry constraint for the depth completion task. It is based on an observation that a generic depth fields can be approximated by a set of low dimensional principle component bases that extracted from natural depth maps. We then assume the recovered dense depth map should be a weighted sum of these bases that satisfy the input sparse depth points. Moreover, in order to deal with the over-smooth problem of the recovered depth maps, we further introduce an auto-regression model to allow the accompanied colour images to guide the reconstruction process. Our method has closed form solutions that can be efficiently solve within several seconds. Moreover, since we do not make any assumptions to the modality of the input, our method can be applied to any kind of depth sensors. Extensive experiments on KITTI dataset and Middlebury dataset show that our method is robust to the sparsity of the input and have good cross dataset performance. Moreover, our method can work well even when the input depth points are corrupted.

\section{Background}
Depth completion task can be roughly classified into two categories: conventional methods and deep learning based methods. Here we only review the most related methods. 

\subsection{Conventional methods}
Diebel \etal \cite{DiebelT05} presented a Markov Random Fields (MRF) which uses colour information and depth information where available. The underlying assumption is that areas of constant colour are most likely to have constant depths. The work \cite{Andreasson07} follows the same assumption and compared five different interpolation methods with \cite{DiebelT05}. More recently \cite{DolsonBPT10} proposed a flexible method that is able to up-convert the range sensor data to match the image resolution. However, it is not a per-frame algorithm that requires as least 2 successive frames. In order to intensively use the structural correlation between colour and depth pairs in colour guided depth super-resolution, non-local means (NLM) was introduced as a high-order term in regularization \cite{Park2011}. \cite{YangAR2014} swapped the Gaussian kernel in the standard NLM to a bilateral kernel to enhance the structural correlation in colour-depth pairs and proposed an adaptive colour-guided auto-regressive (AR) model that formulates the depth upsampling task as AR prediction error minimization, which owns a closed-form solution. A few approaches employ sparse signal representations for guided upsampling making use of the Wavelet domain \cite{Hawe2011} , learned dictionaries \cite{Li2012} or co-sparse analysis models \cite{Gong2014}. \cite{Barron2016} and \cite{Li2016} leverage edge-aware image smoothing techniques and formulate it as a weighted least squares problem while \cite{Li2016} also applied coarse-to-fine strategy to deal with the very sparse situation. 

\subsection{Deep learning based methods}
Recently, deep learning based end-to-end methods were introduced to image super-resolution \cite{Dong2014} and geometry learning tasks such as monocular depth estimation~\cite{Zhong2018ECCV}, stereo matching~\cite{zhong2017self,Zhong2018ECCV_rnn,zhong2020nipsstereo}, optical flow~\cite{Zhong_2019_CVPR,zhong2020nipsflow}. Song \etal \cite{Song2016} extend it to the problem of depth map super-resolution. Riegler \etal \cite{Riegler2016} proposed a deeper network for depth map super-resolution that makes the training process much faster with better performance. However, the main drawback for adopting deep learning to our task is the irregular pattern that makes the correlation between the depth points and their spatial locations vary from frame to frame. Very recently, Sparsity Invariant CNNs \cite{Uhrig2017THREEDV} has been proposed to handle the sparse and irregular inputs. With the guidance of corresponding colour images, Ma \etal  \cite{Ma2018SparseToDense} extended the up-projection blocks proposed by \cite{laina2016deeper} as decoding layers to achieve full depth reconstruction. Jaritz \etal \cite{jaritz2018sparse} presented a method to handle sparse inputs of various densities without any additional mask input. It soon been applied to LiDAR-Stereo Fusion tasks~\cite{Cheng_2019_CVPR}.  Rather than training an end-to-end neural network to perform depth upsampling, several algorithms leverage deep neural networks to provide high-level cues in improving the performance, i.e., pixel-level semantic cues. With very similar objective as our task, Schneider \etal \cite{semantic2016} utilizes semantic input provided by fully convolutional networks (FCNs) \cite{Long2015} and achieves relatively good results. However, since their algorithm relies on semantic segmentation, their performance is largely depended on the performance of segmentation results. 

\section{Problem statement}
We define a general depth completion task as follows. Let us define a set of 3D points that measured from a depth sensor as $\mathbf{X}=\{\mathbf{x}_1, \mathbf{x}_2,...,\mathbf{x}_n\}$, where $\mathbf{x}_i = (x_i,y_i,z_i,1)^T$ is the 3D coordinate of the $i^{th}$ point in homogeneous coordinate. $\mathbf{P}$ is the $3\times 4$ camera projection matrix, which projects 3D points in the world coordinate to 2D image coordinates on the image plane. $\mathbf{T}$ is a $4\times 4$ external matrix that maps 3D points from the sensor's local frame coordinates to world coordinates. The depth measurements $\mathbf{S} = \{\mathbf{s}_1, \mathbf{s}_2, \cdots, \mathbf{s}_n\}$ on the image plane can be derived as $\mathbf{S} = \mathbf{P}  \mathbf{T} \mathbf{X}$, where $\mathbf{s}_i = (u_i,v_i,d_i)^T$, $(u_i,v_i)$ is the $i^{th}$ point's corresponding 2D image coordinate and $d_i$ represents its depth value. The target of this task is to assign a proper depth $d_j$ for each image point $(u_j,v_j)$.

Besides the input sparse depth measurements, we could potentially have other information such as the corresponding high-resolution colour image. With the help of the colour guidance, we expect to predict more accurate depth value $d_j$ for each pixel $(u_j, v_j)$ on the image plane. 

%==========================================================
\section{Method}
%We propose a model that constructs the weight of each pixel by both local correlation in the initial depth map and nonlocal similarity in the accompanied high quality colour image and takes trained PCA components as a global constrain. The advantage of this model is that when the colour-depth pairs have strong correlation, it can accurately recover the shape of object; when it is not, it reliefs the texture copy effects with the help of prior depth knowledge.
In this section, we propose two depth completion methods. We first formulate the depth maps as a weighted sum of PCA bases and try to solve the problem in a closed form solution. Then we add a colour guided auto-regression model to the formulations to boost the performance. 

We use the following priors on depth maps and their accompanied colour images:
\begin{enumerate}
\item The depth maps and the colour images have strong local correlations, \ie the depth map could be expressed/predicted by the colour image in local region;
\item The depth map lies in a low dimensional subspace, \ie each depth map could be represented as a linear combination of basis depth maps.
\end{enumerate}

\subsection{Learning Depth Map Bases}
\begin{figure}[t]
\begin{center} 
\subfigure[PCAs from KITTI]{
\includegraphics[width=0.47\columnwidth,height=2.4cm]{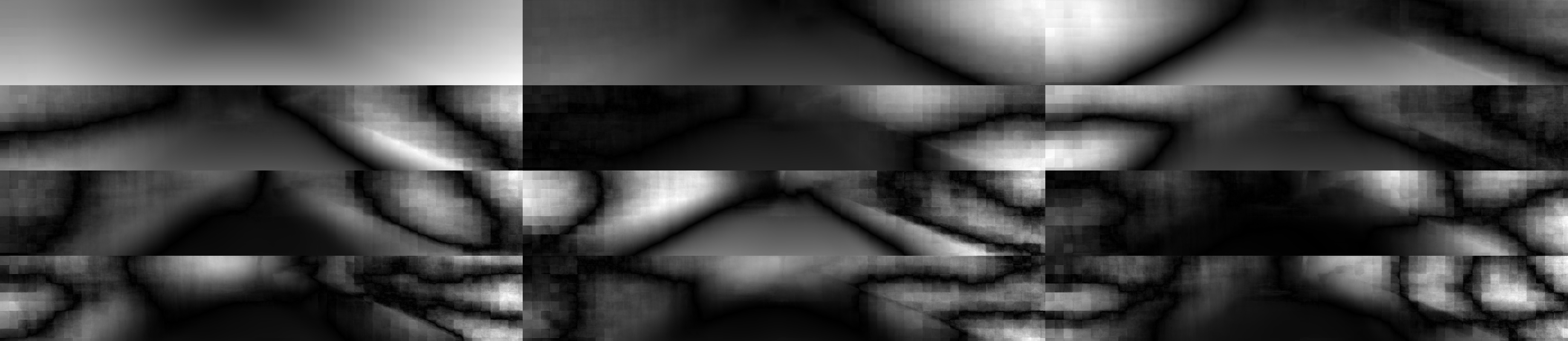} } \hspace{-.2cm}
\subfigure[PCAs from NYU]{
\includegraphics[width=0.47\columnwidth,height=2.4cm]{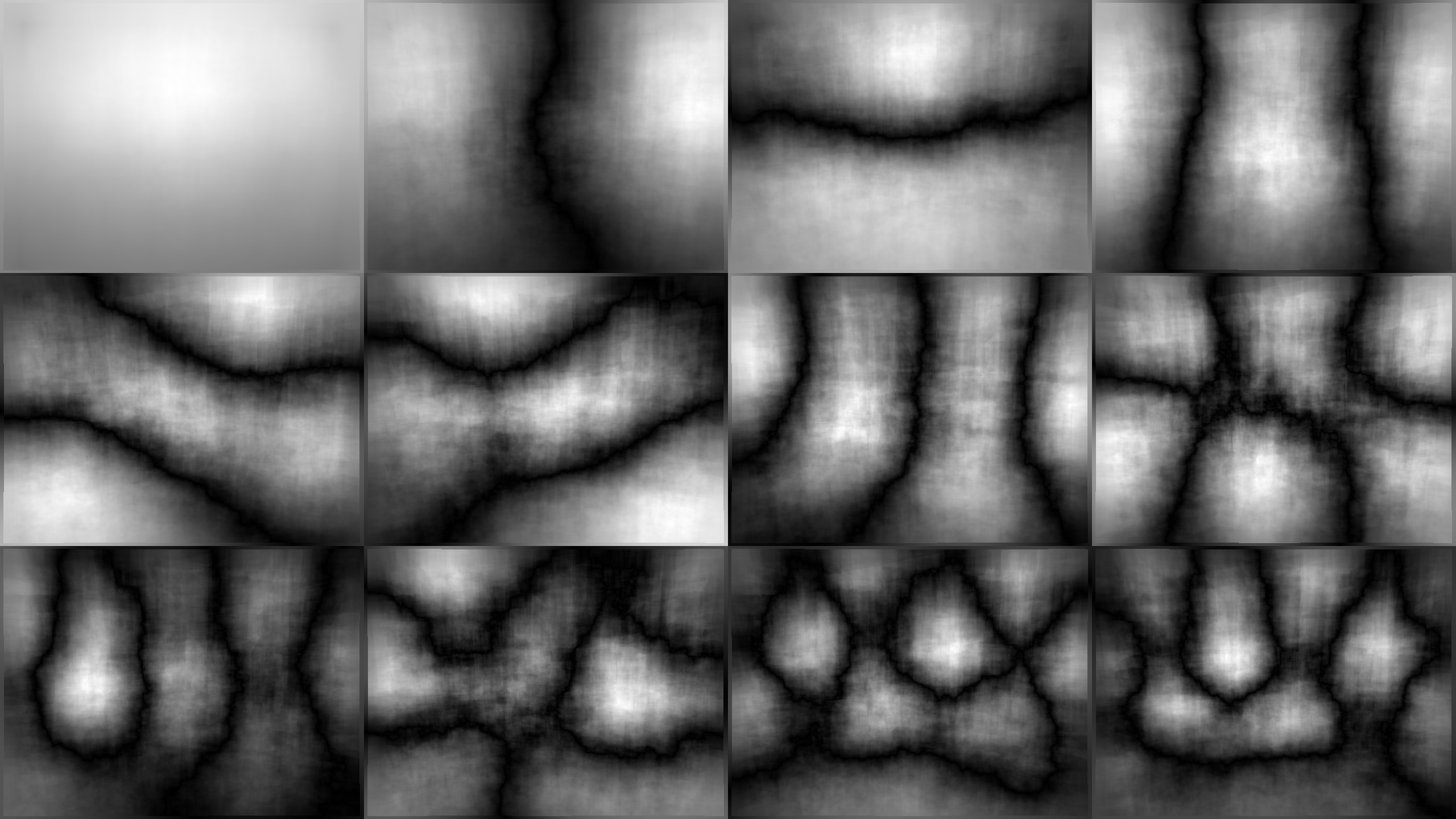} }  
\caption[Extracted 12 PCA components.]{\label{fig:pca}\textbf{The extracted first 12 PCA components.}}
\end{center}
\end{figure}

We extract PCA bases of natural depth maps with the widely used robust PCA \cite{HaubergRPCA2014} algorithm. Note that it is possible to use other basis learning methods or sparse coding methods to generate the bases. Considering the scale differences between outdoor and indoor scenes, we train two set of PCA bases, one from Sequence 10 of KITTI visual odometry dataset (Figure \ref{fig:pca} (a)) for outdoor scenarios and the other from NYUV2 training dataset \cite{Silberman2012} (Figure \ref{fig:pca} (b)) for indoor scene. Since the ground truth depth maps are incomplete, we inpaint the incomplete depth maps \cite{Garcia20101167} before extracting the PCA bases.

\subsection{PCA-based Depth Prediction}
The PCA based depth prediction method consists of the following steps:
\begin{enumerate}[1)]
\item Extract the first $k$ basis $\mathbf{A}$ from pre-existing dense depth maps using robust PCA \cite{HaubergRPCA2014}; 
\item Define the $n$ control points based on depth measurements $\mathbf{X} = \{\mathbf{x}_1,\mathbf{x}_2, \cdots, \mathbf{x}_n\}$, where $\mathbf{x}_i = (u_i,v_i,d_i)^T$;
\item The problem of dense depth prediction now can be formulated as a least squares fitting problem:
\begin{equation}
\mathbf{\widehat{w}} = \arg \min_{\mathbf{w}}\norm{\mathbf{\widehat{A}w}-\mathbf{X}}_2^2
\end{equation}
where $\mathbf{w} = (\mathbf{w_1},\mathbf{w_2},...\mathbf{w_k})^T$ defines the weights and $\mathbf{\widehat{A}}$ is a subset of the PCA bases $\mathbf{A}$ with $n\times k$ dimensions, corresponding to the positions of available measurements. There exists a closed-form solution for the above least squares problem as: 
\begin{equation}
\mathbf{w} = \mathbf{\widehat{A}}^{\dag}\mathbf{X},
\end{equation}
where $\mathbf{\widehat{A}}^{\dag}$ denotes the pseudo-inverse of $\mathbf{\widehat{A}}$.

\item Dense depth $\mathbf{d}$ can be predicted as:
\begin{equation}
\mathbf{d} = \mathbf{A}\mathbf{w}.
\end{equation}
\end{enumerate}
Note that we do not use weighted PCA as we assume all sparse depth measurements are with the same confidence.

\subsection{Colour-guided PCA for Depth Prediction}
Colour-guided smoothness term, aims at representing the local structure of the depth map. Depth maps for generic 3D scenes contain mainly smooth regions separated by curves with sharp boundaries. The key insight behind the colour guided smoothness term is that the depth map and the colour image are locally correlated, thus the local structure of the depth map can be well represented with the guidance of the corresponding high-resolution colour image. The term is widely used in image colourization, depth in-painting and depth image super resolution. We add this term together with the previously proposed global geometry constraint term to further boost the performance.

Denote $D_{u}$ as the depth value at location $u$ in a depth map, the depth map inferred by the model can be expressed as:
\begin{equation}
D_u = \sum_{v \in \theta_{u}} \alpha_{(u,v)} D_v,
\end{equation}
where $\theta_{u}$ is the neighbourhood of pixel $u$ and $\alpha_{(u,v)}$ denotes the colour guided smoothness model coefficient for pixel $v$ in the set of $\theta_{u}$. The discrepancy between the model and the depth map  (the colour guided smoothness potential) can be expressed as: 
\begin{equation}\label{eq:pix}
\psi_{\theta}(\mathbf{D}_\theta) = \left(D_u - \sum_{r\in \theta_{u}}\alpha_{(u,v)}D_v\right)^2.
\end{equation}

We need to design a local colour guided smoothness predictor $\alpha$ with the available colour image. 
\begin{equation}
\alpha_{(u,v)} = \frac{1}{N_{u}} {\alpha^I_{(u,v)}}
\end{equation}
where $N_u$ is the normalization factor, $\overline{D}$ is the observed depth map, $I$ is the corresponding colour image.
\begin{equation}
{\alpha^I_{(u,v)}} = \exp(-\sum_{i\in C}\norm{\mathbf{B_u}\circ (g_u-g_v)}^2_2/(2\times 3\times \sigma^2))
\label{eq:AR}
\end{equation}
where $g$ represents the intensity value of corresponding colour pixels, $\sigma$ is the variance of colour intensities in the local path around $u$. "$\circ$" denote the element-wise multiplication. $\mathbf{B_u}$ represents the bilateral filter kernel: ${\mathbf{B_u}_{(u,v)}} = \exp(-\sum_{i\in C}(g_u-g_v)^2/(2\times 3\times \sigma_{I_u}^2))$, where $\sigma_{I_u}$ controls the importance of intensity difference. The window size for $\theta$ is set as $9\times{9}$ in our experiment. 

With the help of colour-guided auto-regression model \cite{YangAR2014}, we can formulate the problem as a convex minimization with respect to $\mathbf{d}$ and $\mathbf{w}$ simultaneously,
\begin{equation}
\min_{\mathbf{d},\mathbf{w}}\frac{1}{2}\norm{\mathbf{Pd}-\mathbf{d^0}}_2^2+\frac{\lambda}{2}\norm{\mathbf{Qd}-\mathbf{d}}_2^2+\frac{\gamma}{2}\norm{\mathbf{Aw}-\mathbf{d}}_2^2
\label{pcacolour}
\end{equation}
where $\mathbf{P}$ defines the observation matrix, and $\mathbf{Q}$ is a prediction matrix corresponding to colour guided smoothness predictors $\alpha_{(u,v)}$.  The above optimization problem also owns a closed-form solution.

Let's denote $\mathbf{P'} = [\mathbf{P}, \mathbf{0}], (\mathbf{I}-\mathbf{Q})' = [\mathbf{I}-\mathbf{Q},\mathbf{0}], \mathbf{A'} = [\mathbf{I},\mathbf{-A}], \mathbf{x} = [\mathbf{d},\mathbf{w}]$, then the above equation \eqref{pcacolour} can be expressed as:
\begin{equation}
\arg \min_{\mathbf{x}}\frac{1}{2}\norm{\mathbf{P'x}-\mathbf{d^0}}_2^2+\frac{\lambda}{2}\norm{(\mathbf{I}-\mathbf{Q})'\mathbf{x}}_2^2+\frac{\gamma}{2}\norm{\mathbf{A'x}}_2^2.
\label{pcacolour1}
\end{equation}

The closed-form solution to the above optimization is achieved as:
\begin{equation}
(\mathbf{P'}^T\mathbf{P'}+\lambda(\mathbf{I}-\mathbf{Q})'^T(\mathbf{I}-\mathbf{Q})'+\gamma \mathbf{A'}^T\mathbf{A'})\mathbf{x} = \mathbf{P'}^T\mathbf{d}^0
\label{eq.final}
\end{equation}

%===========================================================
\section{Evaluation}
We evaluate our method extensively on well known KITTI stereo and visual odometry (VO) datasets and Middlebury datasets with a comparison to current state-of-the-art approaches.

\subsection{Error Metrics}
We deploy two quantitative measurements: 
\begin{enumerate}[1)]
\item \textbf{Mean relative error} (MRE), which is defined as $\mathrm{e}_{rel} = \frac{1}{N}\sum_{i=1}^N\frac{|d_i-\widehat{d_i}|}{d_i}$, where $d_i$ and $\widehat{d_i}$ are the ground truth depth and depth prediction respectively. A lower MRE indicates a better dense depth prediction performance achieved.

\item \textbf{Bad pixel ratio} (BPR) measures the percentage of erroneous positions in total, where a depth prediction result is determined as erroneous if the absolute depth prediction error is beyond a given threshold $d_{th}$. A lower BPR indicates a better depth prediction results.
\end{enumerate}

BPR and MRE measure different statistics of the dense depth prediction results, which jointly evaluate the prediction performance.

\subsection{Results on the KITTI dataset.}
\begin{figure}
\begin{center} 
%\centering
 % 2
 \subfigure[The input: downsampled Lidar points]{
\includegraphics[width=0.47\columnwidth]{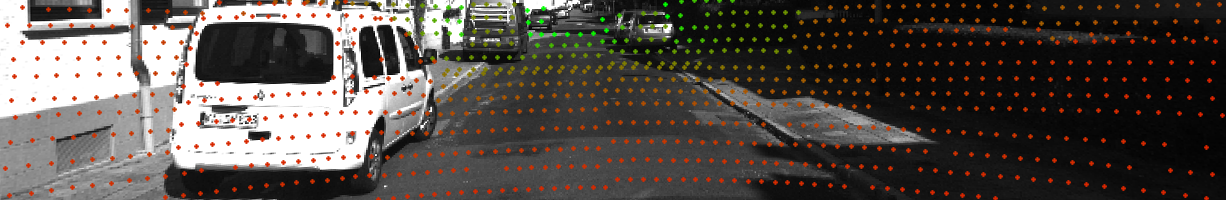}}
 \subfigure[Ground truth depth map provided by the 64-line Velodyne HDL-64E LIDAR.]{
   \includegraphics[width=0.47\columnwidth]{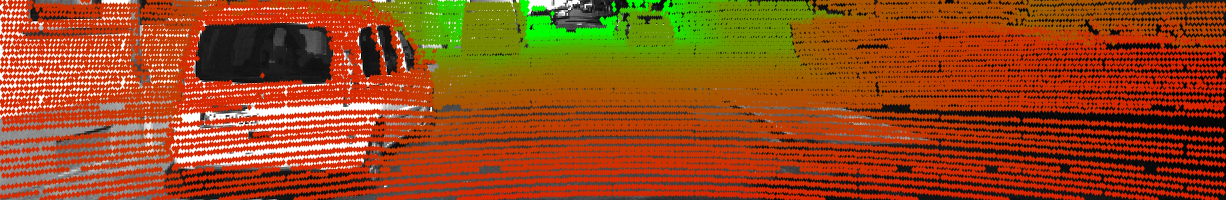}}
 \caption[Experiment setting]{\label{fig:KITTIVOLidar}\textbf{Experiment setting:} Taking sparse and noisy depth measurement in (a) as input, with the learned depth map bases and the guidance of colour image, our method outputs a dense depth map, which is evaluated against the ground truth depth map illustrated in (b).}
 \end{center}
\end{figure}

We perform evaluation of two subset of KITTI dataset: KITTI stereo and KITTI VO, which both consist of challenging and varied road scene imagery collected from a test vehicle. Ground truth depth maps are obtained from 64-line LIDAR data. The main difference between these two is that the ground truth depth maps from the stereo dataset were manually corrected and interpolated based on their neighborhood frames. Note we only used the lower half part of the images ($200\times 1226$) as the upper half part generally include large part of sky and there is no depth measurements available. We down-sampled ground truth depth map with stride of 8 as the input. Our algorithm will recover a dense depth map but we only evaluate its performance on the original sparse depth points. An example of input and ground truth depth map is given in Figure \ref{fig:KITTIVOLidar}. 

In Table \ref{my-label1} and  Table \ref{my-label2}, we compare our method (colour Guide PCA) to state-of-the-art colour guided depth interpolation approach \cite{YangAR2014} on the KITTI 2012 and 2015 datasets, respectively. In Bad Pixel Ratio, we use threshold of 3 meters. Our method outperforms current state-of-the-art results on both dataset with a notable margin. Our method achieves $7.10$ and $8.63$ performance leap for KITTI 2012 and 2015 datasets respectively.

\begin{table}
\centering
\caption[Evaluation on the KITTI stereo 2012 dataset]{\textbf{Evaluation on the KITTI stereo 2012 dataset}}
\label{my-label1}
\tabcolsep=0.55cm
\begin{tabularx}{\columnwidth}{c|c|c|c}
\hline
Percentage      & PCA     & AR \cite{YangAR2014}    & Colour Guide PCA        \\ \hline
Relative Error  & 5.66    & 10.43                   & \textbf{3.33}          \\ \hline
Bad Pixel Ratio & 18.59   & 13.15                   & \textbf{10.19}        \\ \hline
\end{tabularx}
\end{table}

\begin{table}
\centering
\caption[Evaluation on the KITTI stereo 2015 dataset]{\textbf{Evaluation on the KITTI stereo 2015 dataset}}
\label{my-label2}
\tabcolsep=0.55cm
\begin{tabularx}{\columnwidth}{c|c|c|c}
\hline
Percentage      & PCA     & AR \cite{YangAR2014}  & Colour Guide PCA    \\ \hline
Relative Error  & 7.50    & 13.53                 & \textbf{4.90}      \\ \hline
Bad Pixel Ratio & 18.48   & 14.71                 & \textbf{11.84}     \\ \hline
\end{tabularx}
\end{table}

We also perform the quantitative and qualitative evaluation of our method on KITTI VO dataset. KITTI VO dataset consists of 22 sequences 43,596 frames which including vary driving scenarios such as highway, city and country road.  In our experiment setting, we only use the left images and Lidar points. Note, we recover the depth every 10 frames, 4,359 frames in total. 

\begin{figure}[!t]
\begin{center} 
%\centering
 % 2
 \subfigure{
\includegraphics[width=0.47\columnwidth]{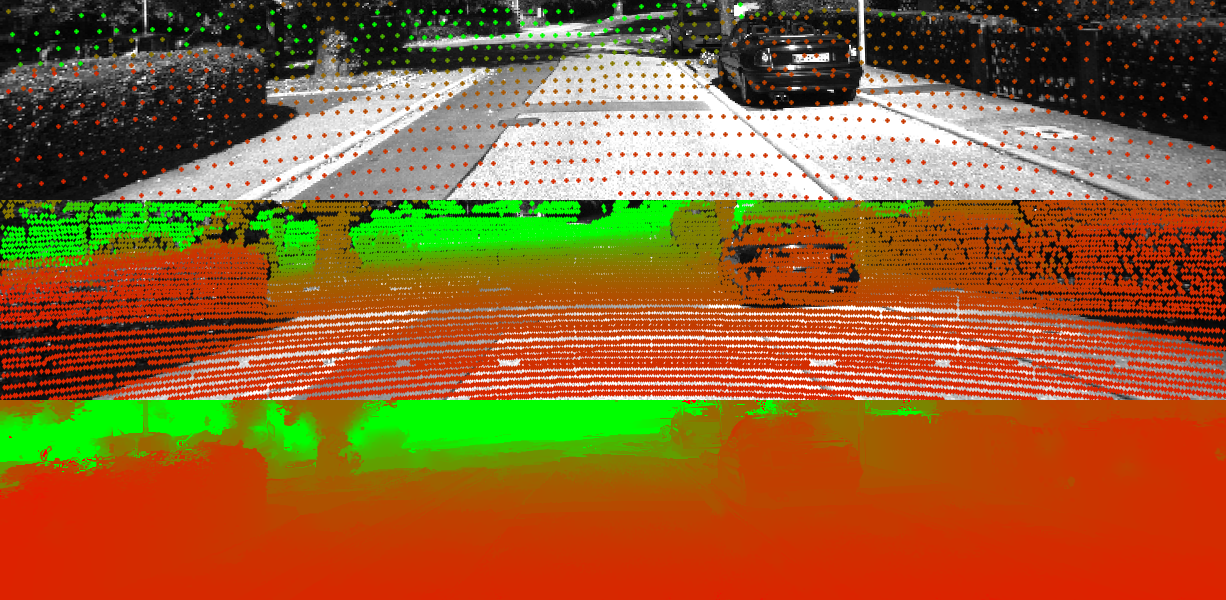} } \hspace{-.31cm}
 % 3
 \subfigure{
   \includegraphics[width=0.47\columnwidth]{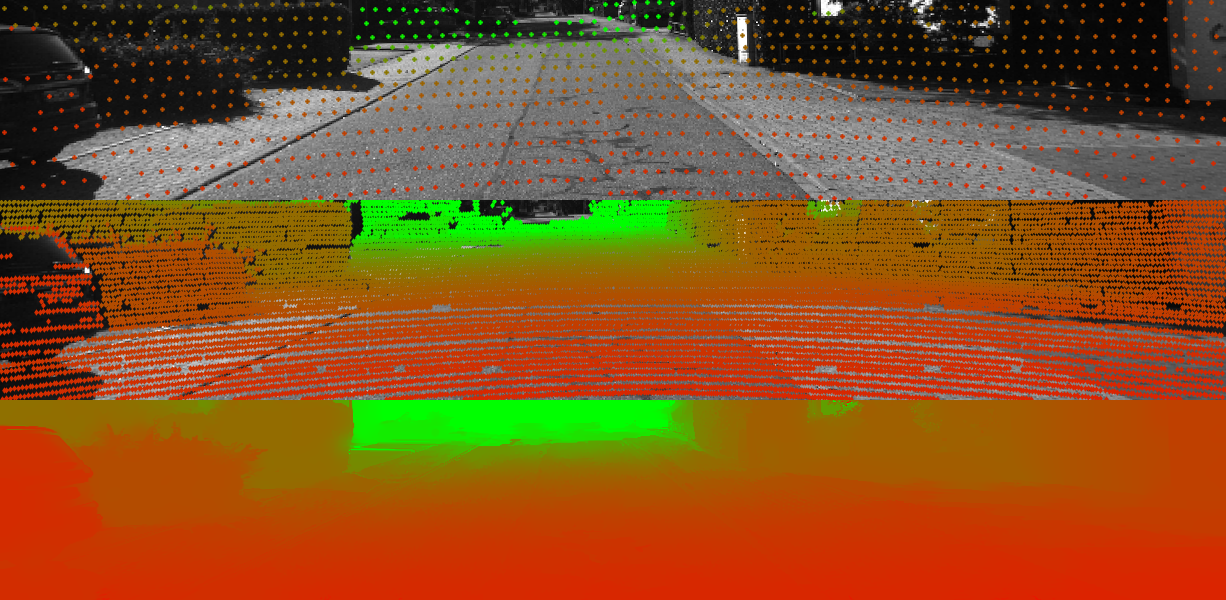} }
 \caption[Recovery results on KITTI VO dataset.]{\label{fig:KITTIVO} \textbf{Recovery results on KITTI VO dataset.} top: the input sparse Lidar points; middle: the ground truth Lidar points; bottom: the recovered dense depth map.}
 \end{center}
\end{figure}

Figure \ref{fig:KITTIVO} illustrates two sample results. Note that our method even successfully recovers the correct depth on car windows which lacks the depth measurement in ground truth depth map. 

\begin{figure}[!t]
\begin{center} 
%\centering
 % 2
 \subfigure[Mean relative error on each sequence.]{
\includegraphics[width=0.48\columnwidth]{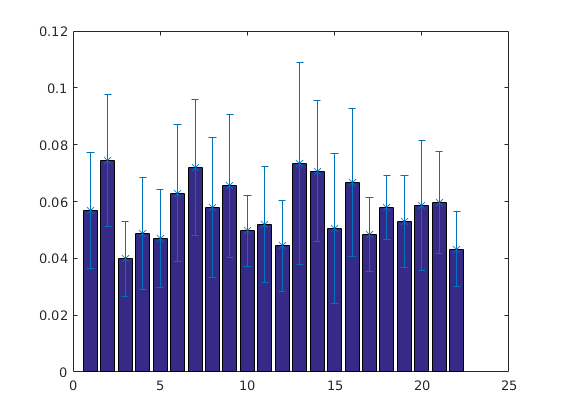} } \hspace{-.31cm}
 % 3
 \subfigure[Bad pixel ratio on each sequence with threshold $3$ meters]{
   \includegraphics[width=0.48\columnwidth]{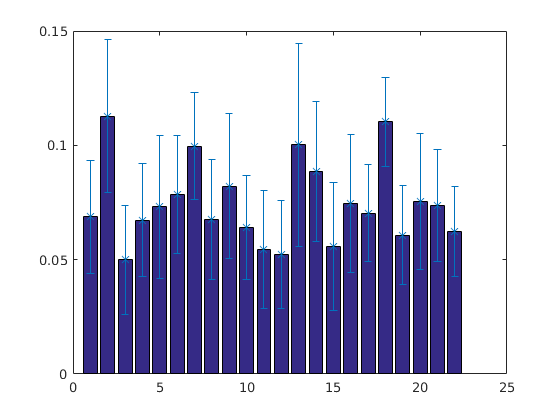} }
 \caption[Quantitative results on each sequences.]{\label{fig:KITTIVOseq} \textbf{Quantitative results on each sequences.} The error bar shows the standard deviation.}
 \end{center}
\end{figure}

\begin{figure}[!t]
\begin{center} 
%\centering
 % 2
 \subfigure[Histogram of Mean relative error on KITTI VO dataset]{
\includegraphics[width=0.45\columnwidth]{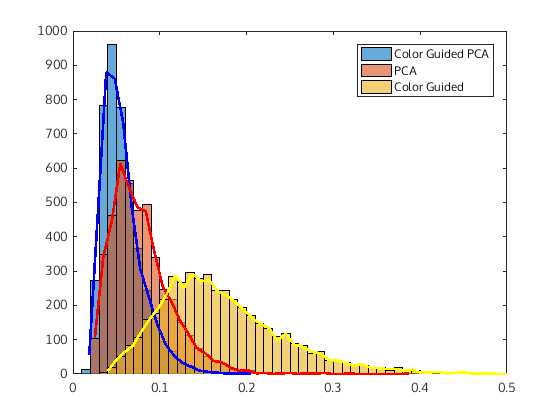} } %\hspace{-.31cm}
 % 3
 \subfigure[Histogram of Bad pixel ratio on KITTI VO dataset]{
   \includegraphics[width=0.45\columnwidth]{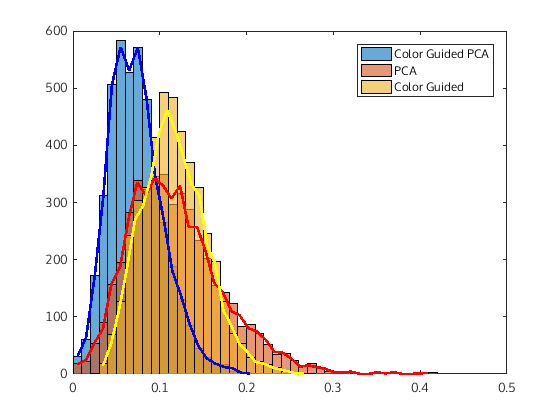} }
 \caption[Quantitative comparison between PCA and colour guidance methods.]{\label{fig:KITTIVOall} \textbf{Quantitative comparison with PCA and colour guidance methods.} Our method notably outperforms all of them.}
 \end{center}
\end{figure}

The quantitative results are shown by sequence in Figure \ref{fig:KITTIVOseq} and by histogram in Figure \ref{fig:KITTIVOall}. The overall Mean relative error is $0.0569$ and the Bad pixel ratio is $0.0725$ with a threshold of 3 meters. Figure \ref{fig:KITTIVOall} also provides the comparison between PCA and colour guided method. Our method performs notable better through out the whole KITTI VO dataset.

\subsection{Results on the Middlebury dataset.}
Our proposed method can also be used as a standard depth upsampling method. We conduct evaluation on the Middlebury dataset with $8\times$ upsampling rate. The resolution of colour guide and ground truth disparity map is set to $640\times 480$. In quantitative comparison, as there are holes in ground truth disparity map, we exclude these holes areas in comparison. Note we use threshold of 1 pixel in Bad Pixel Ratio metric.

Quantitative comparison with state-of-the-art method \cite{ferstl2013b} is shown in Table \ref{middleburymre} and \ref{middleburyd_err} with Mean relative error and Bad pixel ratio respectively. Our method significantly outperform the state-of-the-art in all aspects with average margin $31.21\%$ in MRE and $69.67\%$ in BPR.

\begin{figure}[!t]
\begin{center} 
\subfigure{
\includegraphics[width=0.22\columnwidth]{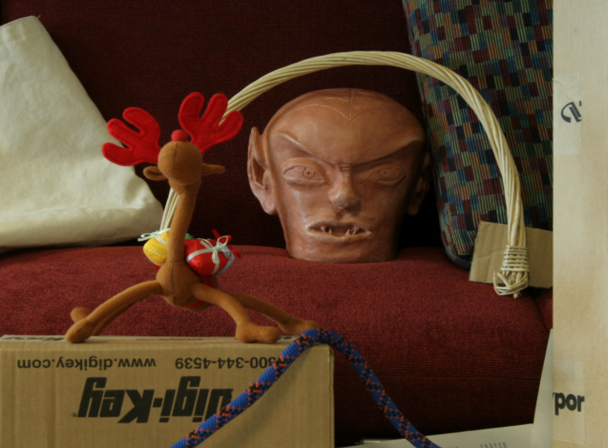} }
\subfigure{
\includegraphics[width=0.22\columnwidth]{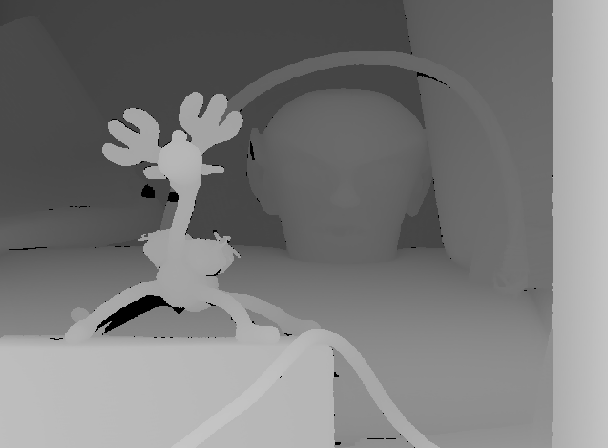} }
\subfigure{
\includegraphics[width=0.22\columnwidth]{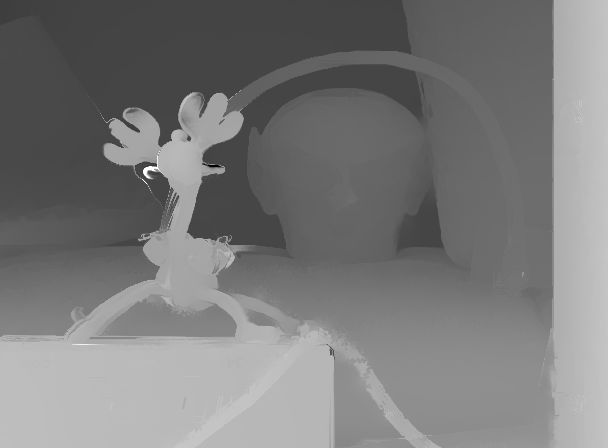} }
\subfigure{
\includegraphics[width=0.22\columnwidth]{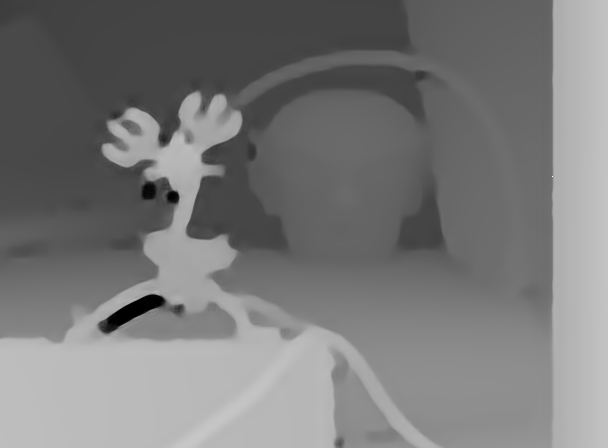} } 
\subfigure{
\includegraphics[width=0.22\columnwidth]{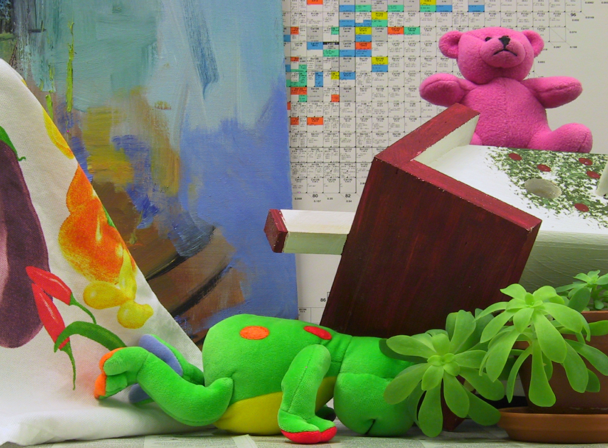} } 
\subfigure{
\includegraphics[width=0.22\columnwidth]{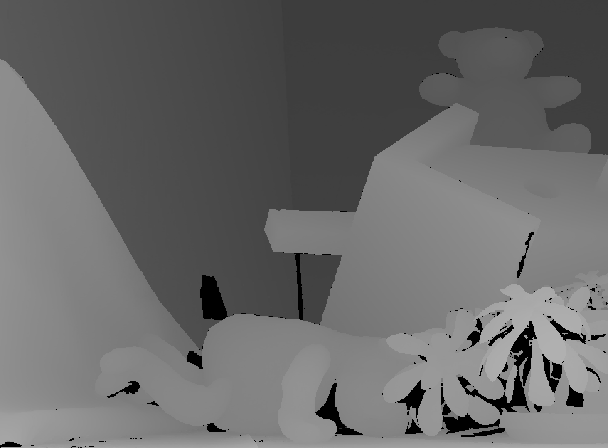} }
\subfigure{
\includegraphics[width=0.22\columnwidth]{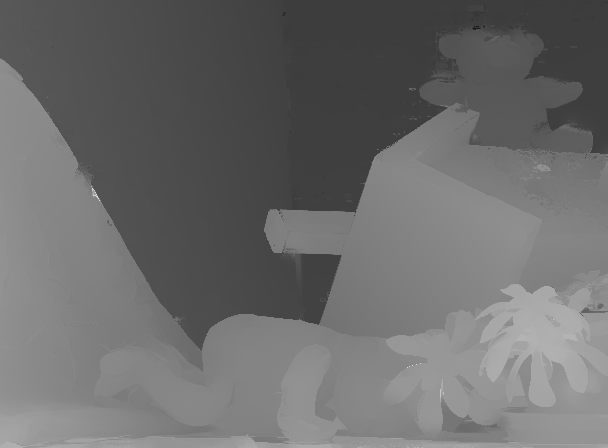} } 
 \subfigure{
   \includegraphics[width=0.22\columnwidth]{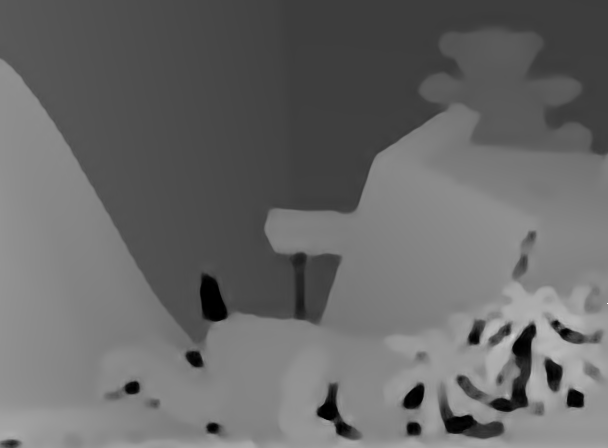} }
 \caption[Recovery results on Middlebury dataset.]{\label{fig:mid} \textbf{Recovery results on Middlebury dataset.} From left to right: colour images, ground truth depth maps, our results and results from \cite{Song2016}.}
 \end{center}
\end{figure}

Figure \ref{fig:mid} illustrates two sample results. Our method produce much sharper and more accurate boundaries. The texture on background board and the words in front board do not misguide our algorithm to generate texture copy effects on the recovered depth map.

\begin{table}[htp]
\centering
\caption[Evaluation in MRE on the Middlebury dataset.]{\textbf{Evaluation in MRE on the Middlebury dataset}}
\label{middleburymre}
\tabcolsep=0.25cm
\begin{tabularx}{\columnwidth}{c|c|c|c|c|c|c}
\hline
Percentage      		& Art  & Books & Dolls & Laundry & Moebius & Reindeer   \\ \hline
Bicubic         		& 3.69 & 1.42  & 1.69  & 2.20    & 1.82    & 2.13       \\ \hline
Bilinear        		& 3.63 & 1.28  & 1.53  & 2.00    & 1.64    & 1.96       \\ \hline
Ferstl\cite{ferstl2013b}& 2.76 & 1.07  & 1.28  & 1.93    & 1.43    & 1.56      \\ \hline
Our method      		& \textbf{2.25} & \textbf{0.98}  & \textbf{0.92}  & \textbf{1.35}    & \textbf{1.03}    & \textbf{1.16}      \\ \hline
\end{tabularx}
\end{table}

\begin{table}[htp]
\centering
\caption[Evaluation in Bad pixel ratio on the Middlebury dataset]{\textbf{Evaluation in Bad pixel ratio on the Middlebury dataset}}
\label{middleburyd_err}
\tabcolsep=0.25cm
\begin{tabularx}{\columnwidth}{c|c|c|c|c|c|c}
\hline
Percentage      		& Art   & Books & Dolls & Laundry& Moebius & Reindeer    \\ \hline
Bicubic         		& 25.42 & 9.93  & 13.15 & 15.73  & 12.73   & 13.39      \\ \hline
Bilinear        		& 20.90 & 8.41  & 12.44 & 13.98  & 11.82   & 11.01     \\ \hline
Ferstl\cite{ferstl2013b}& 16.98 & 11.19 & 14.31 & 15.76  & 12.66   & 11.51     \\ \hline
Our method      		& \textbf{12.75} & \textbf{7.49}  & \textbf{7.83}  & \textbf{7.33}   & \textbf{7.73}    & \textbf{6.62}       \\ \hline
\end{tabularx}
\end{table}

\section{Conclusion}
In this paper, we have proposed a novel PCA-based colour guide depth completion algorithm which is able to recover high quality high resolution depth map from sparse inputs in a closed form solution. Experiments on ``KITTI'' and ``Middlebury'' dataset demonstrate that our algorithm has high resistance on texture copy effects and significantly outperform state-of-the-art methods with sharper boundaries, including deep learning based methods.

%===========================================================
\bibliographystyle{unsrt}
\bibliography{egbib}

%this would normally be the end of your paper, but you may also have an appendix
%within the given limit of number of pages
\end{document}